\documentclass{article}

\PassOptionsToPackage{numbers, compress}{natbib}
\usepackage{graphicx}
\usepackage{amsmath}
\usepackage{biblatex} %Imports biblatex package
\usepackage[nonatbib, preprint]{neurips_2022}

\usepackage[utf8]{inputenc} % allow utf-8 input
\usepackage[T1]{fontenc}    % use 8-bit T1 fonts
\usepackage{hyperref}       % hyperlinks
\usepackage{url}            % simple URL typesetting
\usepackage{booktabs}       % professional-quality tables
\usepackage{amsfonts}       % blackboard math symbols
\usepackage{nicefrac}       % compact symbols for 1/2, etc.
\usepackage{microtype}      % microtypography
\usepackage{xcolor}         % colors
\usepackage{multirow}
\usepackage{multicol}

\title{Improving the Diagnosis of Psychiatric Disorders with Self-Supervised Graph State Space Models}

\author{
  \AND Ahmed El Gazzar \\ Amsterdam University Medical Center \\a.g.elgazzar@amsterdamumc.nl \And Rajat Mani Thomas \\ Amsterdam University Medical Center \\rajatthomas@amsterdamumc.nl \And Guido Van Wingen \\ Amsterdam University Medical Center \\guidovanwinge@amsterdamumc.nl
}

\begin{document}

\maketitle

\begin{abstract}
Single subject prediction of brain disorders from neuroimaging data has gained increasing attention in recent years. Yet, for some heterogeneous disorders such as  major depression disorder (MDD) and autism spectrum disorder (ASD), the performance of prediction models on large-scale multi-site datasets remains poor. We present a two-stage framework to improve the diagnosis of heterogeneous psychiatric disorders from resting-state functional magnetic resonance imaging (rs-fMRI). First, we propose a self-supervised mask prediction task on data from healthy individuals that can exploit differences between healthy controls and patients in clinical datasets. Next, we train a supervised classifier on the learned discriminative representations.  To model rs-fMRI data, we develop \textbf{Graph-S4}; an extension to the recently proposed state-space model \textbf{S4} to graph settings where the underlying graph structure is not known in advance. We show that combining the framework and Graph-S4 can significantly improve the diagnostic performance of neuroimaging-based single subject prediction models of MDD and ASD on three open-source multi-center rs-fMRI clinical datasets.\footnote{https://github.com/neurips22/fMRI-GraphS4}
\end{abstract}

\section{Introduction}

Unlike other fields of medicine, psychiatry lacks diagnostic criteria based on validated biomarkers. Research in the previous years have shown that biomarkers can be extracted from resting-state functional magnetic resonance (rs-fMRI) signal of the brain. In efforts to extract such information, researchers have opted for machine learning to map brain functional activity to diagnostic labels provided by clinical assessments of self-reported symptoms according to the diagnostic and statistical manual of mental disorders (DSM-5) \cite{dsm}. Given the challenging nature of fMRI data (high dimensionality, low sample size, and low signal to noise ratio), machine learning in psychiatry requires advanced architectures to  better model the data and improve the classification metrics and provide more interpretable explanations. While there has been varying success depending on the models, datasets and applications, the literature suggests that results for certain phenotypes and psychiatric disorders have been more successful than others, where several multi-site studies consistently report high classification/regression scores and relatively uniform biomarkers. On the favourable end of the spectrum, there are the objective measures such as age, sex, handedness, and some psychiatric disorders such as schizophrenia \cite{arbabshirani2017single, eickhoff2019neuroimaging, cui2018effect}. On the other end of the spectrum are subjective measures such as IQ, neuroticism, and heterogeneous disorders such as MDD and ASD \cite{quaak2021deep, arbabshirani2017single, gallo2021thalamic}. The focus of this work is on these highly heterogeneous disorders, how to improve their classification performance and how to identify biomarkers for these disorders. 

While there has been rapid acceleration in developing deep learning architectures to diagnose psychiatric disorders, there is is still an ongoing debate in the neuroimaging community about the advantage of adopting these models given that several papers have reported comparative performance using linear models and simple kernel methods on hand-crafted features\cite{schulz2020different, he2020deep}. The current explanations for this include: noise in the data that linearize decision boundaries, small samples sizes, the use of hand-engineered features and inter-site heterogeneity.

While we agree with the general sentiment that simple models perform competitively with sophisticated deep models and observe a similar pattern (see Section \ref{baselines}), we argue that a main obstacle bounding the performance of deep models in clinical datasets is the heterogeneity of the disorders and noise in the labels. There exists several diagnostic sub-types for MDD and ASD which has been shown to drive different neural correlates of functional connectivity \cite{lamers2010identifying, van2012data}. The difference in treatment can have a large impact on resting-state brain activity \cite{wang2012amplitude, zhuo2019rise}.  The inter-rater reliability of MDD diagnosis is quite low (kappa  = 0.28) \cite{regier2013dsm} which further reflects the heterogeneity of the disease. Yet, a large body of deep learning methods approach the problem as an end-to-end supervised binary classifcation problem and devote the focus on model design. While this supervised approach can theoretically work as a normal vs abnormal classification problem in large samples where the model can learn an implicit model of what is normal versus what is not, this is not the case for clinical fMRI datasets where the sample sizes are small, the data is noisy and the supervised task is not explicitly designed to exploit patterns of normative behaviour in the data or the deviation from it. 

To address these challenges, we propose a two stage framework of self-supervised learning on data from healthy individuals followed by supervised learning on learned representations. Self-supervised learning has emerged as a powerful tool to leverage unlabeled data to learn informative representations useful in downstream tasks. %%%Self-supervised learning in the context of one-class classification can be viewed either as i) a direct unsupervised anomaly detector with  density-based\cite{zhai2016deep, du2019implicit, nalisnick2019detecting}, reconstruction-based \cite{zong2018deep, pidhorskyi2018generative} and contrastive-based methods \cite{tack2020csi} or ii) a pretraining method to extract discriminative features useful in a downstream task \cite{sohn2020learning, hendrycks2019using, ,li2021cutpaste}.  In this work we follow the later approach while using an anomaly detection score on an independent validation set as measure of how discriminative the self-supervised task is.
A key aspect in designing a self-supervised task on one-class datasets is to find a task that exploits the normative patterns in the data\cite{tack2020csi, sohn2020learning}. Recently, the most popular self-supervised learning approaches rely on contrastive learning of different views of the same data sample to learn features that can differentiate between instances\cite{chen2020simple}. However, instance discrimination does not explicitly capture normative patterns in the data \cite{tack2020csi}. Further, generating different views by augmentation requires semantic understanding of the data and the effect of the augmentation. Overall, augmentations techniques are under-studied on fMRI datasets and applying random augmentations can drastically change the distribution of the data.(See Appendix A.2 for more discussion and results on contrastive learning on fMRI). Rather, a more constructive approach is to design a task that leverages our limited understanding of the behaviour of the brain activity.  To this end we opt to the field of network neuroscience which defines the brain as a set of interconnected networks operating in synchrony and that brain disorders manifest as disturbed inter-network and or intra-network connectivity \cite{sporns2016networks}. Inspired by self-supervised pretext mask prediction tasks in NLP \cite{devlin2018bert, radford2018improving}, we propose to train a saptio-temporal model to predict the temporal activity of brain regions of interest (ROIs) that belong to a specific pre-defined network given the temporal activity of the rest of the brain. The anomaly score can thus be computed as the prediction error. Finding networks that maximizes the difference between prediction error of normal samples versus disease can highlight potential network level biomarkers and improve downstream classification tasks.

To develop spatio-temporal models for fMRI data,  we adopt the recently proposed state-space model \textbf{S4} \cite{s4} and extend it to a graph setting to fit the nature of the data. The \textbf{S4} model can efficiently model long-range interactions which makes it suitable to learn temporal data from fMRI and generate accurate long-range predictions. To adapt it to the network representation of the brain, we augment it with an adaptive graph convolution layer for feature mixing between ROIs.\\
\textbf{Contributions} We propose a new framework for normative modelling via self-supervised prediction of pre-defined brain networks that enables single-subject detection of heterogeneous psychiatric disorders and identify associated network-level biomarkers without requiring any labelled data of the disorder during training. The labels are only used in a validation set to determine the threshold. Further, We extend the S4 model to graph settings where the graph structure is not known and evaluate its performance as a sequence-to-sequence model in the self-supervised setting and as a classifier in the supervised setting. Finally, We compare several open-source machine learning fMRI classification models on three clinical rs-fMRI datasets.

\section{Related Work}

\paragraph{Single-subject phenotype prediction from fMRI} While there is no one standard method of applying machine learning methods on fMRI data, the most common method used is to parcellate the brain into ROIs, compute the correlation matrix between the temporal activity of the ROIS, then flatten the lower triangular matrix and feed it to a linear classifier or kernel-based classifier such as an SVM. With the development of deep learning, researchers have explored different input representations and different model architectures such as RNNs\cite{fan2020deep, el2019hybrid}, 1D CNNs\cite{el2019simple}, 3D CNNs\cite{thomas2020classifying}, Transformers\cite{malkiel2021pre}, GCNs \cite{li2021braingnn, kim2020understanding} and most popular recently are ST-GCNs \cite{gadgil2020spatio,dynamic-fmri, azevedo2020towards, kim2020understanding}. A majority of these works were developed and evaluated on objective objective measures such as sex classification or task classifcation form task-based fMRI and report (80\%-98\%) test accuracies. However, works developed on multi-site MDD and ASD datasets reported far less impressive test metrics. (64\%-72\%) were reported by \cite{heinsfeld2018identification,el2019simple, thomas2020classifying, yan2019reduced} for ASD diagnosis on \textbf{ABIDE} \cite{abide} dataset and (61\%-63\%) were reported by \cite{restmdd,gallo2021thalamic} for MDD diagnosis on \textbf{Rest-Meta-MDD}\cite{restmdd} dataset.

\paragraph{Self-Supervised learning on rs-fMRI} The success of self-supervised learning on other data domains have inspired it's application on fMRI data. There has been a large focus on the task of denoising/reconstruction using auto-encoders.\cite{heinsfeld2018identification} trained an auto-encoder on the correlation vector to provide better initialization. \cite{malkiel2021pre} utilized 3D CNNs and Transformers to train an auto-encoder to reconstruct the 3D structure of the scan. Most close to this work, is the work by \cite{suk2016state} in which they trained an auto-encoder to reconstruct the temporal activity of all the brain ROIs, followed by training  Hidden Markov model (HMMs) on the embeddings to diagnose mild cognitive impairment. 

\paragraph{State-space-models on fMRI} Over the last two decades, the application of SSMs have grown increasingly popular to infer effective connectivity from fMRI data under certain stimuli or to decode the brain dynamic signal at rest into meaningful cognitive states \cite{hutchinson2009modeling, eavani2013unsupervised, janoos2011spatio}. Recently \cite{suk2016state, kusano2019deep} have utilized HMMs with deep models as a probabilistic methods to diagnose psychiatric disorders

\section{Methods}

\subsection{Problem formulation}

Given a small labelled clinical dataset $D_c = \{X_i, Y_i\}_{i=1}^{N_c}$ and a large subset of a population-level dataset with no self-reported symptoms of psychiatric disorders $D_p = \{X_i\}_{i=N_c+1}^{N_c + N_p}$ where $N_p >> N_c$, 
 $X_i$ $\in$ $R^{V \times T}$ is a  resting-state brain functional where $V$ is the number of brain regions as defined by a brain atlas and  $T$ is the number of sampled data-points for the duration of the scan.  $Y_i$ is a binary clinical diagnosis given by a clinician (or a vote of clinicians) for a specific disorder based on self-reported symptoms of participant $i$. Our objective is to learn a predictive model on the clinical $f$ $\colon$ $X $  $\mapsto$  $Y$. Given the limited sample size of the clinical dataset, the heterogeneity of the disorder and the noise in the data and labels,  solely training $f$ by supervised learning is likely to overfit the training samples. We propose to leverage the healthy population dataset  $D_p$ and a healthy subset of the clinical dataset $D_c$ to learn a normative pretext task  $f_{ss}$ $\colon$ $X$ $\mapsto$ $Z$, then utilize the rest of the clinical labelled dataset to learn a supervised model  $f_{sup}$ $\colon$ $Z$ $\mapsto$ $Y$. The proposition of this work is that learning a latent representation $Z$ that can capture differences between normal and abnormal samples at inference time would alleviate the problem of disease heterogeneity at a low sample size during supervised training. The key challenges then become i) how to design $f_{ss}$ to exploit differences between healthy controls and patients? , ii) how to learn \textit{transferable} $f_{ss}$ between the population and clinical datasets given the difference between scanners, acquisition protocol and demographics?, and iii) how to build efficient deep learning models for fMRI that can capture the spatio-temporal dynamics  of the signal?

\subsection{Self-supervised learning with network mask prediction}
The objective of applying self-supervised techniques on one-class datasets is to learn a proxy task on the class distribution that can maximize the discriminative power of the learned representations between in- and out-of-distribution samples. For rs-fMRI, finding such a task is non-trivial given the fine-grained inter- and intra- group differences and the stochastic nature of resting-state brain activity. In this work, we propose to utilize the prior domain knowledge about the functional organization of the brain as a set of distributed networks operating in parallel to design the self-supervised task. Specifically, we group the brain ROIs into 8 networks, 7 cortical networks as defined by \cite{yeo2011organization} and the sub-cortical regions. The task is then to train a sequence-to-sequence model to predict the temporal activity of ROIs belonging to one network given the temporal activity of the ROIs belonging to the other 7 networks. Formally, given a pre-defined brain network $B_i$ $\in$ [$B_1$, ..., $B_8$] with $B$ $\in$ $R^{w \times h \times d}$, we can generate a binary mask $M$ $=$ $(m_{it})$  $\in$ $[0,1]^{V \times T}$ of the same dimension as the input signal $X$, the mask vector $m_i$ of the $i$-th region is defined as:
\[
    m_i = 
\begin{cases}
    1 & \text{if} \text{  ROI$_{i}$  $\cap$ $B$ $\geq$ $0.5$} \\
    0,              & \text{otherwise}
\end{cases}
\]

and the task can be defined as $f_{ss}$ $\colon$ $X$ $\odot$ $M$  $\mapsto$ $X$ $\odot$ $\neg$ $M$. This task is analogous to self-supervised mask prediction in NLP and computer vision applications. However instead of the masked tokens/patches sampled randomly, here the mask is sampled according to pre-defined brain networks and the model has to predict the entire temporal profile of the masked regions. This context to content design exploits dynamic functional connectivity and subsequently learns discriminative features useful in downstream classification tasks. Further, evaluating the target network with the highest discriminative power between patients and controls could be informative of the underlying neurobiology of the disorder. Note that while it is possible to train one model to predict all the networks using uniform network sampling during training, we empirically find that training one model per brain network resulted in better discriminative features and improved downstream classification performance.

\subsection{Self-supervised loss}
 A good time-series prediction $\hat{y}$ should have a low absolute error with respect to the ground truth $y$ and follow its shape. The vast majority of time-series prediction models rely on minimizing the Mean Squared Error (MSE) or it's variants (MAE, RMSE, etc.) between the ground truth and prediction. This is sub-optimal for non-stationary signals as pointed and addressed by \cite{dilate}. While the rs-fMRI temporal signal is relatively stationary with slow oscillation, for some idle brain regions with low amplitude fluctuations, an MSE objective could result in predictions that have a low error representing the mean of the oscillations while ignoring the shape of the signal (See Appendix B.1 for visualizations). To address this limitation, we introduce a simple Pearson's correlation term the MSE objective as follows:

\begin{align}\label{loss}
    L(y,\hat{y}) = \lambda_{1}\frac{1}{V}\frac{1}{T}\sum_{i=1}^{V} \sum_{j=1}^{T} ||y_{ij} - \hat{y}_{ij}||^{2}_{2} - \lambda_{2}
    \frac{1}{V}\sum_{i=1}^{V} \frac{\sum_{j=1}^{T}(y_{ij} - \bar{y}_{j}) (\hat{y}_{ij} - \bar{\hat{y}}_{j}) }
    {\sqrt{(\sum_{j=1}^{T}(y_{ij} - \bar{y}_{j})^{2}} \sqrt{\hat{y}_{ij} - \bar{\hat{y}}_{j})^{2}}}
\end{align}

where $\bar{y}_{j}$and $\bar{\hat{y}}_{j}$ are the scalar mean values of $y_j$ and $\hat{y}_j$ respectively and $\lambda_1$, $\lambda_2$ are hyper-parameters.

The organization of brain networks have been validated and replicated across different studies with different demographics and scanning hardware which further support the generalizability of the task\cite{power2011functional,yeo2011organization}. To address the distributional shift between the population dataset and the clinical dataset, we first train the self-supervised task on the population dataset $D_p$ for the first 20 epochs to provide better initialization, then resume self-supervised training on a subset of healthy data in the clinical dataset $D_{c-ss-train}$. To identify which network prediction task is crucial for discriminating between healthy participants and patients in the clinical dataset, we use an independent balanced validation set $D_{c-ss-val}$ and calculate the masked network prediction MSE as an outlier score. The use of a held out validation set to optimize the self-supervised tasks is common practice in self-supervised learning, where self-supervised training is followed by supervised training of a linear layer on top of the representation and the results on a validation set are used to find the optimal self-supervised task (e.g. which augmentation techniques to use in contrastive learning \cite{chen2020simple}). Note that in our pipeline, no supervised training is required as we can use the prediction error as an anomaly score. Further, as we later present in Section \ref{results}, identifying which network prediction task has the highest discriminative power can provide potential biomarkers for the targeted clinical group.

\subsection{Graph State Space models}

\subsubsection{The S4 model}

The formulated task is a sequence-to-sequence problem that entails learning long-range temporal correlations for the duration of the resting-state scan ($T$ $\approx$ 500 timepoints). While conventional methods such as CNNs, RNNs and  Transformers have specialized variants for capturing long-range dependencies \cite{liprnns,cnnsrnns,informer,wavenet}, training these models on fMRI datasets remains challenging given the limited sample size of the dataset which constrains the number of trainable parameters and/or model depth necessary to capture global context. 
The recently introduced \textbf{S4} model \cite{s4}, utilizes state space models (SSM) as trainable layers for learning from sequential data and show excellent results in various domains and excel in long range tasks.

State space models in continuous time-space maps a 1-D input signal $u(t)$ to an N-D latent state $z(t)$ before projecting to a 1-D output signal $y(t)$ as follows:

\begin{align}
\begin{split}\label{eq:1}
    z^{'}(t) = Az(t) + Bu(t)
\end{split}\\
\begin{split}\label{eq:2}
    y(t) = Cz(t) + Du(t)
\end{split}
\end{align}

To operate on discrete-time sequences sampled with a step size of $\Delta$, the SSM can be discretized using the bilinear method \cite{bilinear} as follows:
\begin{align}
\begin{split}\label{eq:3}
    z_{k} = \bar{A}z_{k-1} + \bar{B}u_{k}  \qquad    y_{k} = \bar{C}z_{k} + \bar{D}u_{k}
\end{split}
\end{align}
\begin{align}
\begin{split}\label{eq:4}
    \bar{A} = (I- \Delta/2 \cdot A)^{-1} (I + \Delta/2 \cdot A ) \qquad  \bar{B} = (I- \Delta/2 \cdot A)^{-1} \Delta B ) \qquad \bar{C} = C 
\end{split}
\end{align}
Given an initial state $z_{k} = 0$ and omitting $D$ (as it can be represented as a skip connection in the model), unrolling eq:\ref{eq:3} yields:
\begin{align}
\begin{split}\label{eq:5}
   y_{k} = \bar{C}\bar{A}^k\bar{B}u_{0} + \bar{C}\bar{A}^{k-1}\bar{B}u_{1} + ... + \bar{C}\bar{B}u_{k}\\
  \textit{in vector form }\qquad  y = \bar{K} * u  \qquad \bar{K} = (\bar{C}\bar{A}^i\bar{B})_{i \in [L] }
\end{split}
\end{align}
 
$\bar{K}$ can thus be interpreted as a convolution filter and the state space model can be trained as a black box sequence to sequence model via learning parameters $A$,$B$,$C$ and $\Delta$ via gradient descent. Efficient learning of $\bar{K}$ requires several computational tricks such as parametrization of $A$ as a diagonal plus low-rank (DPLR) matrix. This parameterization has two key properties. First, this is a structured representation
that allows faster computation using the Cauchy-kernel algorithm \cite{cauchy1,cauchy2} to compute the convolution kernel $K$ very
quickly. Second, this parameterization includes certain special matrices called HiPPO matrices \cite{hippo}, which
theoretically and empirically allow the SSM to capture long-range dependencies better. For in-depth details of the model we refer the readers to \cite{annotateds4}.

SSMs defines a map from $R^L$ $\mapsto$ $R^L$, i.e. a 1-D sequence map. To handle multi-dimensional inputs/features  $R^{L \times H}$, The S4 model simply defines $H$ independent copies of itself at each each layer and after applying a non-linear activation function,  the $H$ feature maps are mixed with a position-wise linear layer. This is analogous to depthwise-separable convolutions \cite{howard2017mobilenets} where independent convolution filters are applied to each feature map independently followed by a linear (1x1) convolution across the depth dimension. 

\subsubsection{Graph-S4}

The S4 model provides a powerful tool for temporal modelling of the brain functional signal and specifically for the task of long time sequence prediction. However, applying S4 off-the-shelf on fMRI data can be highly inefficient. First, S4 layers do not incorporate any spatial inductive bias about the data. A fully connected layer is shared across all the timepoints to mix the independent features. Furthermore, independent states and state-transition matrices for each ROI/feature map is inefficient in modelling the brain signal given the high correlation nature of functional activity between brain ROIs and the limited number of data samples.
To address these limitations, we model the data $X$ as a signal on a graph  $\{G = (V,E)\}$, where $V$ are graph nodes and $E$ $\in$ $R^\{V \times V\}$ represent an undirected dense adjacency matrix initialized randomly from a uniform distribution. Given a $R^V$ dimensional input, instead of learning $V$ state transition matrices $A$ per model layer,  we propose to use a shared transition matrix between the ROIs/features per model layer. Using the depth-wise separable convolution analogy, we use a convolutional kernel with the same weights across the depth dimension.

For feature mixing, we propose to model the mixing process as a diffusion process between graph nodes. \cite{li2017diffusion} have proposed a diffusion convolution layer
which proves to be effective in spatial-temporal modeling. Diffusion convolutional has been further used as graph convolutional layer in several spatio-temporal graph neural network architectures \cite{gwnet, wu2020connecting, dynamic-fmri}. A diffusion convolution layer with input $X$ and output $Z$ can be defined as:

\begin{align}
  Z &= \sum_{d=0}^{D} P^{d}XW_{d}
\end{align}

where $P$ represents the power series of the adjacency matrix of the graph, $D$ is the number of diffusion steps and $W$ is the model weights. For an fMRI graph, defining an adjacency matrix is a non-trivial task \cite{dynamic-fmri} and can be highly dependent on the desired objective of the model. Similar to \cite{dynamic-fmri} we propose an an adaptive adjacency matrix. 

\begin{align}
    E = I_V + \textit{Sparsemax}(\text{ReLU}(E_a \cdot E_a^{T}))
\end{align}

where $E_a$ $\in$ $R^{V \times C}$ is a learnable node embedding dictionaries. Note, that unlike \cite{gwnet, bai2020adaptive, dynamic-fmri} which uses the \textit{Softmax} function to normalize the adjacency matrix, we propose to use \textit{Sparsemax} \cite{martins2016softmax} function as we empirically observe that it provides stable graph structures across different runs and eliminates the need for setting an explicit graph sparsification threshold. The normalized adaptive adjacency matrix can thus be considered as the transition matrix of a hidden diffusion process and the graph convolution layer can defined as 
\begin{align}
  Z &= \sum_{d=0}^{D} EXW_{d}
\end{align}

\section{Experiments}

\subsection{Datasets}

We evaluated our results on 3 different multi-site rs-fMRI clinical datasets \\
1) \textit{\textbf{Rest-Meta-MDD}} \cite{restmdd} is currently the largest open-source rs-fMRI database for studying major depressive disorder including clinically diagnosed patients and healthy controls from 25 cohorts in China with $N$ $=$ 1453 (628 HC/825 MDD).\\
2) \textit{\textbf{SRPBS}} \cite{japanmdd} is a multi-disorder rs-fMRI dataset from 7 cohorts in Japan. We used the open-source subset of the dataset with $N$ $=$ 449 (264 HC/185 MDD) from 4 independent sites. \\
3) \textit{\textbf{ABIDE I+II}} \cite{abide} contains a collection of rs-fMRI brain images aggregated across 29 institutions. It includes data from participants with autism spectrum disorders and typically developing participants (TD). In this study, we used a subset of the dataset with $N$ $=$ 1207 (558 TD/649 ASD).\\

For the population dataset, we utilized the large-scale database of \textit{\textbf{UKBioBank}} \cite{ukbiobank} and sub-sampled a dataset ($N$ $=$ 19000) who did not self-report any symptoms of depression for the two weeks prior to the time of scanning or any lifetime clinical diagnosis of a psychiatric disorder. 

For all the datasets, we adopted the \textbf{Harvard Oxford} \cite{ho} atlas to segment the brain into 118 ROIs (97 cortical/ 21 sub-cortical) and \textbf{Yeo 7-networks} to further group the cortical regions into 7 networks + 1 network that encompasses the 21 sub-cortical regions. \textit{For details about the pre-processing pipeline, sample composition and subject selection criteria, please see the Appendix C}

\subsection{Model Architecture \& Training}

We used a fixed model architecture for all of our self-supervised experiments. Number of Graph-S4 layers $S$ $=$ 4, hidden state dimension = 128, number of channels $C$ $=$ 5, graph diffusion steps $K$ $=$ 2, \textit{GELU} activation function \cite{gelu} as a non-linearity between layers and layer dropout $=$ 0.2.
We trained the models using a batch size of 128 with AdamW optimizer\cite{adamw} at an initial learning rate of  0.01 (with exponential decay) to minimize eq. \ref{loss} with $\lambda_{1}$ and $\lambda_{2}$ as hyper-parameters. We train the models for 20 epochs on the UkBiobank dataset and continued for 100 epochs on the clinical dataset with an early stop of 5 epochs after no improvement on the validation loss on an inner-validation split. Training the model until convergence on the \textit{\textbf{UKBioBank}} followed by the clinical dataset takes approximately 40 minutes on a 16GB Nvidia P100 GPU.

\subsection{Self-supervised learning results}\label{results}

\paragraph{Graph-S4 outperforms other models in predicting brain networks}

We compared the performance of Graph-S4 to other popular long-range sequence-to-sequence models such as \textit{S4} \cite{s4}, \textit{Informer} \cite{informer}, \textit{Wavenet} \cite{wavenet}, \textit{GwNet} \cite{gwnet}. For conciseness, we report here the results on the \textbf{UkBioBank} under a 5-fold cross validation scheme. For a comparison against the baseline models on the clinical datasets, see Appendix D)
The results in Table \ref{table1} showcase the superiority of the \textit{Graph-s4} model against the baseline models on most of the network prediction tasks. The under-performance  of the \textit{Informer} model compared to the rest can be attributed to overfitting on the training data where the model is typically trained on orders-of-magnitude larger datasets. The second best performing model is the \textit{S4} model which further demonstrates the model capacity to capture long-range correlations in comparison to the other models as demonstrated on several forecasting tasks in \cite{s4} and audio generation in \cite{sashimi}.

% \usepackage{multirow}
\begin{table}
\caption{Mean 5-fold MSE of Graph-S4 against several sequence-to-sequence baselines in predicting brain networks on the UKBioBank dataset. (\textit{See Appendix E.1 for implementation details of the baselines.})}\label{table1}
\centering
\begin{tabular}{|l|llllllll|l|}
\hline
\multirow{2}{*}{Model}         & \multicolumn{8}{c|}{\textit{Brain Networks}}                                                                                                                                                                                                                                            &                \\
                               & \multicolumn{1}{l|}{Default}       & \multicolumn{1}{l|}{Vis.}        & \multicolumn{1}{l|}{Somato.}       & \multicolumn{1}{l|}{Vent.}       & \multicolumn{1}{l|}{Dor.}        & \multicolumn{1}{l|}{Front.}         & \multicolumn{1}{l|}{Limb.}        & Sub.          & \textit{Avg.}  \\ \hline
\textit{Wavenet} \cite{wavenet}                        & \multicolumn{1}{l|}{0.25}          & \multicolumn{1}{l|}{0.3}           & \multicolumn{1}{l|}{\textbf{0.16}} & \multicolumn{1}{l|}{0.25}          & \multicolumn{1}{l|}{0.23}          & \multicolumn{1}{l|}{0.11}          & \multicolumn{1}{l|}{0.60}          & 0.51          & 0.301          \\ \hline
\textit{Informer} \cite{informer}                       & \multicolumn{1}{l|}{0.31}          & \multicolumn{1}{l|}{0.44}          & \multicolumn{1}{l|}{0.29}          & \multicolumn{1}{l|}{0.34}          & \multicolumn{1}{l|}{0.31}          & \multicolumn{1}{l|}{0.17}          & \multicolumn{1}{l|}{0.65}          & 0.57          & 0.385          \\ \hline
\textit{GwNet} \cite{gwnet}                         & \multicolumn{1}{l|}{0.23}          & \multicolumn{1}{l|}{0.33}          & \multicolumn{1}{l|}{0.17}          & \multicolumn{1}{l|}{0.25}          & \multicolumn{1}{l|}{0.22}          & \multicolumn{1}{l|}{0.11}          & \multicolumn{1}{l|}{0.59}          & 0.47          & 0.296          \\ \hline
\textit{S4} \cite{s4}                             & \multicolumn{1}{l|}{\textbf{0.18}} & \multicolumn{1}{l|}{\textbf{0.27}} & \multicolumn{1}{l|}{0.19}          & \multicolumn{1}{l|}{\textbf{0.19}} & \multicolumn{1}{l|}{0.22}          & \multicolumn{1}{l|}{0.10}          & \multicolumn{1}{l|}{0.59}          & 0.43          & 0.271          \\ \hline
\textit{Graph-S4} & \multicolumn{1}{l|}{0.22}          & \multicolumn{1}{l|}{\textbf{0.27}} & \multicolumn{1}{l|}{0.17}          & \multicolumn{1}{l|}{\textbf{0.19}} & \multicolumn{1}{l|}{\textbf{0.20}} & \multicolumn{1}{l|}{\textbf{0.09}} & \multicolumn{1}{l|}{\textbf{0.55}} & \textbf{0.39} & \textbf{0.260} \\ \hline
\end{tabular}\\

\end{table}

\paragraph{Self-supervised network mask prediction as an anomaly detector for psychiatric disorders}

In this section we highlight that predicting certain networks can act as weak anomaly detectors for specific disorders. After self-supervised training on the population dataset followed by a healthy subset of the clinical dataset, we calculated the MSE error of the masked network prediction task on an independent balanced validation subset of the clinical dataset. We report the AUROC on the validation set for each of the 8 networks on the three clinical datasets using the Graph-S4 model in Table \ref{table2}. On both depression datasets, predicting the default mode network from the rest brain resulted in  the highest discrimination power followed by the sub-cortical prediction with AUROCs of 0.63 and 0.59 for the \textbf{Rest-Meta-MDD} dataset and 0.65 and 0.6 for the \textbf{SRPBS} dataset, while predicting the rest of the networks did not provide any discriminatory information with AUROCs $\leq$ 0.55. For the ABIDE dataset, only predicting the sub-cortical network provided discrimination between healthy controls and patients with an AUROC $=$ 0.68. This points towards abnormal inter- or intra-network default mode network and sub-cortical connectivity in patients with MDD and abnormal sub-cortical connectivity in patients with ASD. These findings are promising for the potential discovery of network-level biomarkers for psychiatric disorders as it aligns with a large body of the psychiatric neuroimaging literature which reports increased connectivity between the sub-cortical regions and the rest of the brain networks in ASD \cite{cerliani2015increased, sharma2018autism, restingasd, hong2019atypical}, and abnormal default mode network connectivity in depression \cite{yan2019reduced, hamilton2015depressive, grimm2009altered, kaiser2015large}. The MDD literature also points towards  hyper-connectivity of sub-cortical regions (e.g. the thalamus) to the rest of the brain \cite{gallo2021thalamic, greicius2007resting} which could explain the sub-cortical network prediction results in the both MDD datasets. 

% \usepackage{multirow}
\begin{table}
\caption{AUROC for the classification of patients vs controls on the three clinical datasets using the MSE network prediction error with Graph-S4.}\label{table2}
\centering
\begin{tabular}{|l|llllllll|}
\hline
\multirow{2}{*}{Dataset} & \multicolumn{8}{c|}{\textit{Brain Networks}}                                                                                                                                                                                                      \\
                         & \multicolumn{1}{l|}{Default}       & \multicolumn{1}{l|}{Visual} & \multicolumn{1}{l|}{Somato.} & \multicolumn{1}{l|}{Ventral} & \multicolumn{1}{l|}{Dorsal} & \multicolumn{1}{l|}{Front.} & \multicolumn{1}{l|}{Limbic} & Sub.          \\ \hline
\textit{\textbf{Rest-Meta-MDD}}           & \multicolumn{1}{l|}{\textbf{0.63}} & \multicolumn{1}{l|}{0.54}   & \multicolumn{1}{l|}{0.54}    & \multicolumn{1}{l|}{0.50}    & \multicolumn{1}{l|}{0.55}   & \multicolumn{1}{l|}{0.50}   & \multicolumn{1}{l|}{0.44}   & 0.59          \\ \hline
\textit{\textbf{SRPBS}}                   & \multicolumn{1}{l|}{\textbf{0.65}} & \multicolumn{1}{l|}{0.50}   & \multicolumn{1}{l|}{0.56}    & \multicolumn{1}{l|}{0.48}    & \multicolumn{1}{l|}{0.47}   & \multicolumn{1}{l|}{0.39}   & \multicolumn{1}{l|}{0.47}   & 0.60          \\ \hline
\textit{\textbf{ABIDE I+II}}               & \multicolumn{1}{l|}{0.52}          & \multicolumn{1}{l|}{0.54}   & \multicolumn{1}{l|}{0.48}    & \multicolumn{1}{l|}{0.52}    & \multicolumn{1}{l|}{0.46}   & \multicolumn{1}{l|}{0.51}   & \multicolumn{1}{l|}{0.49}   & \textbf{0.68} \\ \hline
\end{tabular}\\

\end{table}

\paragraph{Other self-supervised tasks}

We explored the application of other self-supervised tasks as anomaly detectors on the validation set using Graph-S4. Namely, we experiment with the following tasks. \textbf{Forecast-30}: forecast the temporal activity of all ROIs for the last 30 timepoints using the last $T-30$ timepoints. \textbf{Forecast-50} : forecast the temporal activity of all ROIs for the last 50 timepoints using the last $T-50$ timepoints. \textbf{Denoising} : Recover the input signal after applying random Gaussian noise on the input. \textbf{Random Masks}: The same as network mask prediction but the masked regions are selected randomly. At inference time, we sample 5 random masks and compute the mean MSE as an anomaly score.

We train the tasks using the same training settings as described in Section 4.2. The results at Table \ref{table3} show that while the model can solve these tasks with low errors, it fails to detect out-of-distribution classes. We attribute this to the nature of the tasks that do not exploit inter-and intra network connectivity. Put simply, a well-optimized model can forecast the relatively stationary signal of fMRI with ease or exploit the high correlation between brain regions in denoising and random mask prediction (See Appendix B.2 for visualizations).

\begin{table}
\caption{MSE for self-supervised task and AUROC for the classification of patients vs controls on validation subsets of the three clinical datasets with Graph-S4.}\label{table3}
\centering
\begin{tabular}{|l|ll|ll|ll|}
\hline
\multirow{2}{*}{Task} & \multicolumn{2}{l|}{\textit{\textbf{Rest-Meta-MDD}}} & \multicolumn{2}{l|}{\textit{\textbf{SRPBS}}} & \multicolumn{2}{l|}{\textit{\textbf{ABIDE I+II}}} \\ \cline{2-7} 
                      & MSE         & AUC                  & MSE     & AUC              & MSE        & AUC                \\ \hline
Forecast-30           & 0.16        & 0.49                 & 0.13    & 0.45             & 0.20       & 0.47               \\ \hline
Forecast-50           & 0.33        & 0.53                 & 0.40    & 0.55             & 0.37       & 0.52               \\ \hline
Denoising             & 0.07        & 0.45                 & 0.09    & 0.44             & 0.11       & 0.47               \\ \hline
Random Masks          & 0.13        & 0.54                 & 0.17    & 0.55             & 0.22       & 0.49               \\ \hline
Predict-Default       & 0.28        & \textbf{0.63}        & 0.23    & \textbf{0.65}    & 0.19       & 0.52               \\ \hline
Predict-Subcortical   & 0.51        & 0.59                 & 0.47    & 0.60             & 0.44       & \textbf{0.68}      \\ \hline 
\end{tabular}\\

\end{table}

\subsection{Supervised fine-tuning}
Here, we show the results of supervised fine-tuning with the diagnostic labels of the clinical datasets. Specifically, we froze the weights of the self-supervised pretrained model (including the adaptive adjacency matrix) except the last Graph-s4 layer. We attached a classification head consisting of a global average pooling across the temporal dimension and a linear classification layer with output dimension equal to 2. We trained the classification head and the last graph-s4 layer with cross entropy loss using an AdamW optimizer with a learning rate of 0.001 and a batch size of 128 for 50 epochs with early stop based on the balanced accuracy of an inner validation set. We report the results based on a repeated (10 times) 5-fold cross-validation scheme on the clinical dataset excluding $D_{c-ss-val}$ from the entire process and excluding $D_{c-ss-train}$ from the validation folds. 
We show the results using different model backbones and self-supervised tasks in Table \ref{table4}. Note that for the other self-supervised tasks, we empirically find that freezing the early layers of the model resulted in inferior performance over full fine-tuning of the model. We report the best results and the supervised training strategy for each model-task combination.

\begin{table}
\caption{5-fold balanced accuracy \% (mean $\pm$ sd.) with supervised training using different model/self-supervised task combinations on the the three clinical datasets. }\label{table4}
\centering
\begin{tabular}{|l|l|l|l|l|l|}
\hline
Model    & \multicolumn{1}{c|}{\begin{tabular}[c]{@{}c@{}}Self-Supervised\\ Task\end{tabular}} & \begin{tabular}[c]{@{}l@{}}Supervised\\ Fine-tuning\end{tabular} & \textit{\textbf{Rest-Meta-MDD}} & \textit{\textbf{SRPBS}} & \textit{\textbf{ABIDE I+II}} \\ \hline
Informer & Forecast-50                                                                         & all                                                              & 58.6 $\pm$ 4                    & 62.2 $\pm$ 4            & 64.0 $\pm$ 3                 \\ \hline
Wavenet  & Predict-Default                                                                     & last                                                             & 64.9 $\pm$ 2                    & 69.8 $\pm$ 2            & 57.3 $\pm$ 3                 \\ \hline
GWnet    & Random Masks                                                                        & all                                                              & 62.7 $\pm$ 3                    & 64.4 $\pm$ 3            & 71.9 $\pm$ 2                 \\ \hline
S4       & None                                                                                & -                                                                & 62.9 $\pm$ 2                    & 61.1 $\pm$ 3            & 70.4 $\pm$ 2                 \\
         & Predict-Default                                                                     & last                                                             & 66.2 $\pm$ 2                    & \textbf{70.7 $\pm$ 2}   & 58.2 $\pm$ 4                 \\
         & Predict-Subcortical                                                                 & last                                                             & 61.7 $\pm$ 2                    & 64.8 $\pm$ 3            & \textbf{74.9 $\pm$ 1}        \\ \hline
Graph-S4 & None                                                                                & -                                                                & 63.1 $\pm$ 2                    & 63.7 $\pm$ 4            & 70.3 $\pm$ 1                 \\
         & Denoising                                                                           & all                                                              & 63.2  $\pm$ 2                   & 66.1 $\pm$ 2            & 71.5 $\pm$ 2                 \\
         & Forecast-50                                                                         & all                                                              & 64.1 $\pm$ 1                    & 66.3 $\pm$ 3            & 71.2 $\pm$ 2                 \\
         & Predict-Default                                                                     & last                                                             & \textbf{68.7 $\pm$ 2}           & \textbf{71.4 $\pm$ 2}   & 58.8 $\pm$ 3                 \\
         & Predict-Subcortical                                                                 & last                                                             & 62.9 $\pm$ 3                    & 64.2 $\pm$ 2            & \textbf{76.3 $\pm$ 2}        \\ \hline
\end{tabular}
\end{table}

\subsection{Current supervised fMRI classifiers}\label{baselines}
In Table \ref{table5}, we present the performance of the supervised \textit{Graph-S4} model without any pretraining against existing supervised fMRI classifiers with publicly available code. The methods can be grouped into two distinct groups according to their input data representation. 1) \textit{Static}  where the brain scan is represented as the Pearson's correlation matrix between brain ROIs. 2)\textit{Dynamic} where the input contains the temporal profile of brain ROIs. The results without pre-training show comparative performance between all the models including simple kernel based models such as support vector machines. We do not observe significant improvement either from adopting more advanced models or incorporating dynamic information. This suggest that the methods are bounded by the datasets and in supervised learning settings, more statistically powerful methods can not fulfill their potential.

% \usepackage{multirow}
\begin{table}
\caption{5-fold mean test metrics of several supervised fMRI models on the three clinical datasets.The top entries of the table represent static models and the bottom entries represent dynamic models. \textit{(For details about the models see Appendix E.2)}}\label{table5}
\centering
\begin{tabular}{|l|lll|lll|lll|}
\hline
\multicolumn{1}{|c|}{\multirow{2}{*}{Model}} & \multicolumn{3}{c|}{\textit{\textbf{Rest-Meta-MDD}}}              & \multicolumn{3}{c|}{\textit{\textbf{SRPBS}}}                     & \multicolumn{3}{c|}{\textit{\textbf{ABIDE I+II}}}                \\ \cline{2-10} 
\multicolumn{1}{|c|}{}                       & Acc.                           & Sens.  & Spec. & Acc.                           & Sens. & Spec. & Acc.                           & Sens. & Spec. \\ \hline
\textbf{SVM-linear}         & 60.8                           & 68.1   & 53.5  & 59.4                           & 60.3  & 58.5  & 67.7                           & 75.0  & 60.4  \\
\textbf{SVM-rbf}            & 61.9                           & 73.0   & 50.9  & 60.7                           & 62.1  & 58.8  & 69.5                           & 75.9  & 63.0  \\
\textbf{BrainNetCNN} \cite{kawahara2017brainnetcnn}       & 58.4                           & 50.1   & 56.8  & 60.5                           & 63.9  & 57.1  & 66.6                           & 72.1  & 61.2  \\
\textbf{GIN} \cite{kim2020understanding}               & 62.7                           & 67.3   & 58.1  & 61.8                           & 59.4  & 62.2  & 68.5                           & 81.7  & 55.4  \\ \hline
\textbf{ST-GCN} \cite{gadgil2020spatio}            & 58.2                           & 48.6   & 67.8  & 60.9                           & 60.3  & 61.5  & 65.3                           & 67.2  & 63.4  \\
\textbf{DAST-GCN}  \cite{dynamic-fmri}         & 60.7                           & 44.6   & 75.8  & 60.6                           & 63.7  & 57.5  & 67.8                           & 70.8  & 64.9  \\
\textbf{1D-CNN}  \cite{el2019simple}           & \textbf{63.8} & 65.7 3 & 61.9  & \textbf{63.4} & 65.4  & 61.3  & \textbf{70.6} & 72.7  & 68.5  \\
\textbf{S4}   \cite{s4}              & 62.9                           & 67.5   & 60.1  & 61.1                           & 62.0  & 60.2  & 70.4                           & 74.5  & 66.1  \\
\textbf{Graph-S4}           & 63.1                           & 65.2   & 61.4  & \textbf{63.7} & 62.0  & 60.2  & 70.3                           & 73.4  & 67.2  \\ \hline
\end{tabular}\\

\end{table}

\section{Discussion}
This work highlights disease heterogeneity as one of the main challenges hindering neuroimaging-based predictive models of psychiatric disorders such as MDD and ASD.  Coupled with limited sample sizes and noise in rs-fMRI, we argue and empirically show that end-to-end supervised learning on current clinical datasets does not cut it, even if more sophisticated models are used. Inspired by the success of the self-supervised learning and the emergence of large-scale population datasets such as the UkBioBank, we propose a self-supervised task on data from healthy individuals that exploits dynamic functional connectivity of brain networks to learn disease-discriminative representations. We show that using the task prediction error as an outlier detector on an independent validation set guide towards network-level biomarkers that align with the literature for both disorders. Our results show that supervised learning on learned representations can improve the diagnosis accuracy by 5.6, 7.7 and 6.0 accuracy points for the three clinical dataset used in this study. To develop learning models for fMRI data, we proposed \textit{Graph-S4} ; a simple extension of the \textit{S4} model to graph-structured data. We hope that this work can inspire more knowledge-guided research efforts in self-supervised learning of the brain dynamics. Since scaling up clinical rs-fMRI datasets is not practically feasible, we ultimately believe that self-supervised normative models have promising potential in expanding our understanding of brain function and dysfunction.

\paragraph{Limitations} This work has a number of limitations. The self-supervised task is limited by pre-defined networks and ROIs which is still a method of feature engineering that might hinder the learning process. Further, the adaptive graph structure is static, which is not precisely truthful to the underlying behaviour of the data. Finally, we did not explicitly account for inter-site heterogeneity which might have further hindered the results. Although site harmonization is an active research field, there is currently no agreement on a solution for dynamic timecourses. 

\paragraph{Societal Impact} The advancement of self-supervised methods on population datasets can alleviate the burden of collecting larger clinical datasets. Normative modelling has the potential to uncover novel sub-types of diseases and to advance precision psychiatry. Negative societal impacts can arise if the model is used in the clinic for diagnosis without more validation by certified psychiatrists. 

\printbibliography 
%%%%%%%%%%%%%%%%%%%%%%%%%%%%%%%%%%%%%%%%%%%%%%%%%%%%%%%%%%%%

%%%%%%%%%%%%%%%%%%%%%%%%%%%%%%%%%%%%%%%%%%%%%%%%%%%%%%%%%%%%
\section*{Checklist}

\begin{enumerate}

\item For all authors...
\begin{enumerate}
  \item Do the main claims made in the abstract and introduction accurately reflect the paper's contributions and scope?
    \answerYes
  \item Did you describe the limitations of your work?
    \answerYes
  \item Did you discuss any potential negative societal impacts of your work?
   \answerYes
  \item Have you read the ethics review guidelines and ensured that your paper conforms to them?
    \answerYes
\end{enumerate}

\item If you are including theoretical results...
\begin{enumerate}
  \item Did you state the full set of assumptions of all theoretical results?
    \answerNA{}
        \item Did you include complete proofs of all theoretical results?
    \answerNA{}
\end{enumerate}

\item If you ran experiments...
\begin{enumerate}
  \item Did you include the code, data, and instructions needed to reproduce the main experimental results (either in the supplemental material or as a URL)?
    \answerYes
  \item Did you specify all the training details (e.g., data splits, hyperparameters, how they were chosen)?
    \answerYes
        \item Did you report error bars (e.g., with respect to the random seed after running experiments multiple times)?
    \answerYes 
        \item Did you include the total amount of compute and the type of resources used (e.g., type of GPUs, internal cluster, or cloud provider)?
    \answerYes
\end{enumerate}

\item If you are using existing assets (e.g., code, data, models) or curating/releasing new assets...
\begin{enumerate}
  \item If your work uses existing assets, did you cite the creators?
    \answerYes
  \item Did you mention the license of the assets?
    \answerYes
  \item Did you include any new assets either in the supplemental material or as a URL?
   \answerYes
  \item Did you discuss whether and how consent was obtained from people whose data you're using/curating?
    \answerYes
  \item Did you discuss whether the data you are using/curating contains personally identifiable information or offensive content?
    \answerYes
\end{enumerate}

\item If you used crowdsourcing or conducted research with human subjects...
\begin{enumerate}
  \item Did you include the full text of instructions given to participants and screenshots, if applicable?
    \answerNA{}
  \item Did you describe any potential participant risks, with links to Institutional Review Board (IRB) approvals, if applicable?
    \answerYes
  \item Did you include the estimated hourly wage paid to participants and the total amount spent on participant compensation?
    \answerNA{}
\end{enumerate}

\end{enumerate}

%%%%%%%%%%%%%%%%%%%%%%%%%%%%%%%%%%%%%%%%%%%%%%%%%%%%%%%%%%%%

\newpage
\appendix

\section{Negative Results}

\subsection{One class classification and self-supervised tasks}
The motivation behind this work was to try to find methods to limit the over-reliance on the diagnostic clinical labels in open-source fMRI datasets which groups patients under one label with minimum to no information about disease severity, sub-type, disease severity and medication. Earlier, we opted for a supervised end-to-end one-class classification (OCC) strategy where the model is first trained on the healthy participants from the Ukbiobank dataset followed by fine-tuning on the healthy participants in the clinical dataset. However, all the experiments on the three clinical datasets lead to trivial solutions with the model only predicting the normal class at test time. By looking into the literature, we observed a similar behaviour for end-to-end OCC, where they perform fairly well on simpler datasets (MNIST, COIL100), they fail to generalize to more complex datasets (CIFAR-10, MVTECH-AD, Abnormal 1001). Methods such as \cite{tack2020csi, sohn2020learning, li2021cutpaste} have leveraged data transformations to either generate synthetic outliers or train a pretext self-supervised task useful for downstream one-class classifcation tasks. In contrast to images and videos where we know the casual factors that define the normal/abnormal pattern and can thus engineer transformations to generate in- and out-of-distribution samples for self-supervised training, in fMRI data, we do not have access to this knowledge. Rather, our line of thought was to leverage our limited knowledge of the behaviour of the data to design a pretext task that can learn discriminative representations. At first, we opted for simple tasks such as reconstruction, denoising, forecasting, inter-hemispheric prediction (prediction activity of right hemisphere from left hemisphere or vice versa). The model performance in solving this task on the test data has exceeded our expectation however, this superior performance persisted on both the healthy and patients samples. See Figures \ref{fig2} and \ref{fig3} for a visualization. While training these models we also observed relatively fast convergence and robustness to changing the model hyper-parameters which suggested that these tasks are 'easy' to solve and do not exploit interesting dynamics, rather they rely on the high spatial and temporal correlation in the signal to generate predictions. This was further supported by the lack of discriminative information in the learned representations. One of the co-authors then suggested to look into brain networks and the idea behind was work was born. 

\subsection{Contrastive self-supervised learning}
We examined whether contrastive learning can be useful for rs-fMRI data in learning high-level representations that can improve downstream clinical diagnosis tasks. For dynamic representations of fMRI, we utilized the SimCLR \cite{chen2020simple}  technique and explored different multivariate timeseries augmentations techniques using \textbf{tsaug}\footnote{https://tsaug.readthedocs.io/en/stable/index.html} package in python. For the model encoder, we developed a 1D-CNN encoder similar to \cite{el2019simple} and a MLP with 2 layers for the projection head. We evaluated the quality of the representations using the linear evaluation method \cite{chen2020simple} on an independent validation set. The best performing set of augmentations were resizing the time-course followed by applying Gaussian noise. The balanced accuracies for the \textbf{Rest-Meta-mdd} and \textbf{ABIDE I+II} datasets and with linear evaluation were 0.561 and 0.589 respectively. Supervised fine-tuning of the encoder with MLP classification head resulted did not provide any improvement over the results of the 1D-CNN model on Table \ref{table5}. 
For the static representations of rs-fMRI, we explored the recently proposed \textbf{SCARF}\footnote{\textit{(D Bahri et al. 2022)}} technique for self-supervised learning on tabular data. Specifically it utilizes a similar pipeline to \cite{chen2020simple} but proposes to generate the positive views by randomly corrupting a \% of the features using re-sampling from the marginal distribution of the features. We follow the same model and training parameters used in the paper which was evaluated on 69 real world tabular datasets. Self-supervised pre-training using SCARF resulted in marginal improvement in diagosing psychiatric disorders compared to  training the model from scratch (Balanced accuracies from 0.614 $\pm$ 0.03 to 0.621 $\pm$ 0.02 on \textbf{Rest-Meta-mdd} and  from 0.677 $\pm$ 0.02 to 0.691 $\pm$ 0.01 \textbf{ABIDE I+II}). These results are still comparable with the supervised baselines on Table \ref{table5}.

\newpage

\section{Visualizations}

 \begin{figure}[!h]
\includegraphics[width=\textwidth]{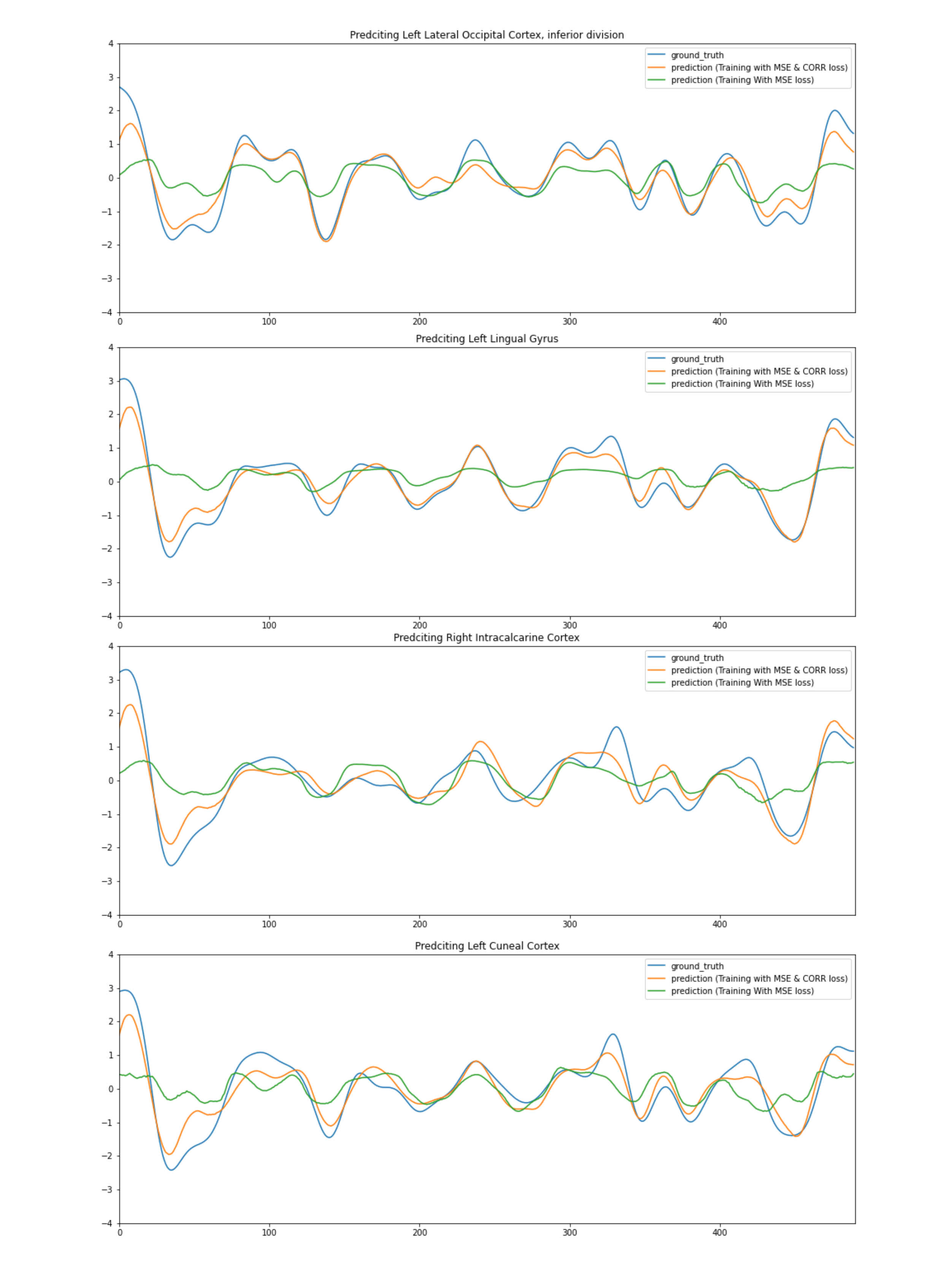}
\caption{Comparison of the Graph-S4 test-set predictions of some ROIs from the visual network when trained using MSE loss only versus the proposed loss function. Note that these samples are selected to highlight the difference. For some ROIs in other brain networks, training using the MSE loss only can be sufficient to generate good predictions.} \label{fig1}
\end{figure}

\begin{figure}[!h]
\includegraphics[width=\textwidth]{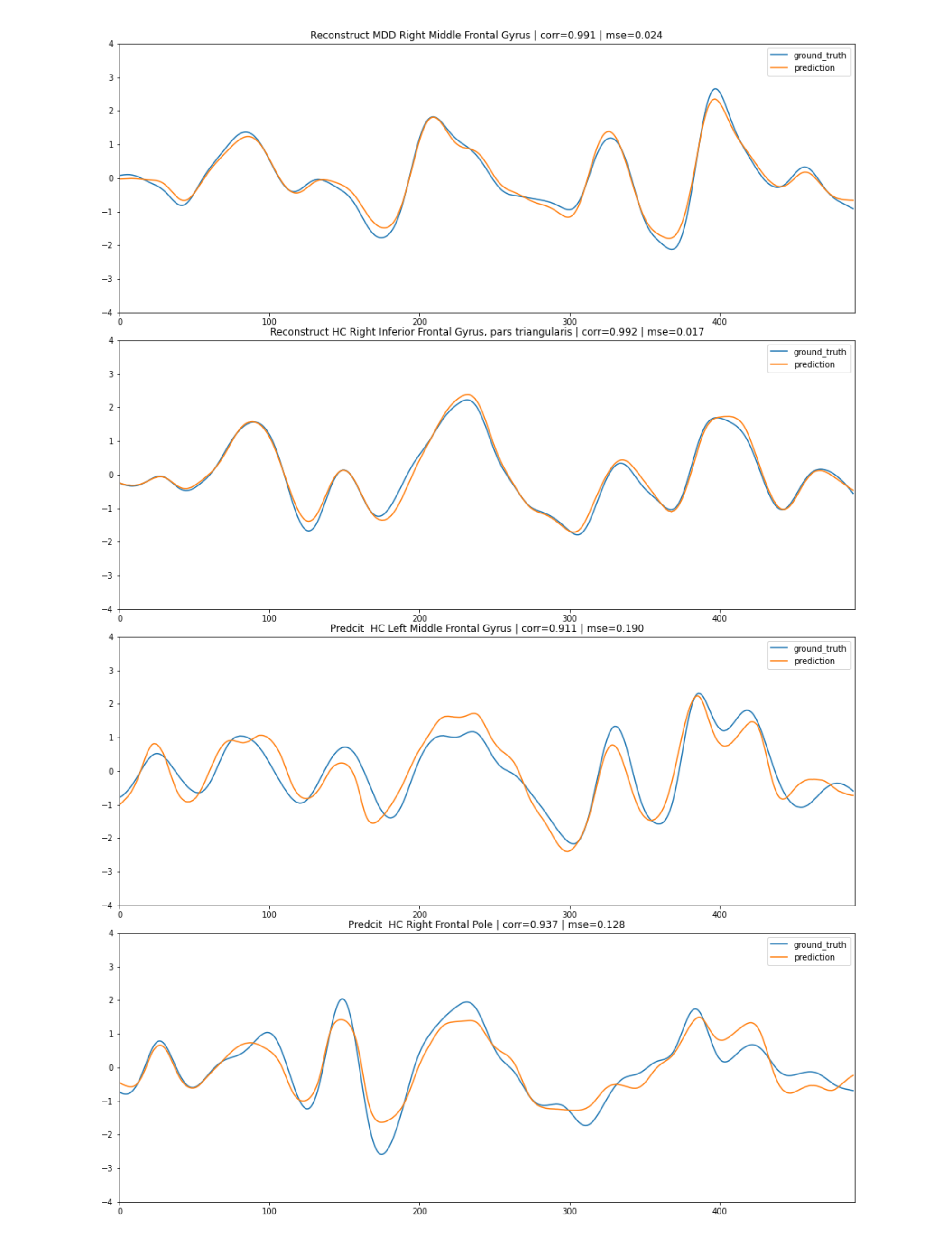}
\caption{Visualization of the Graph-S4 test-set predictions using other self-supervised learning tasks on the healthy participants. \textit{Reconstruct} refers to reconstructing an input with added Gaussian noise. \textit{Predict} refers to predicting random masked regions in the input (no network masking).} \label{fig2}
\end{figure}

\newpage
\begin{figure}[!]
\includegraphics[width=\textwidth]{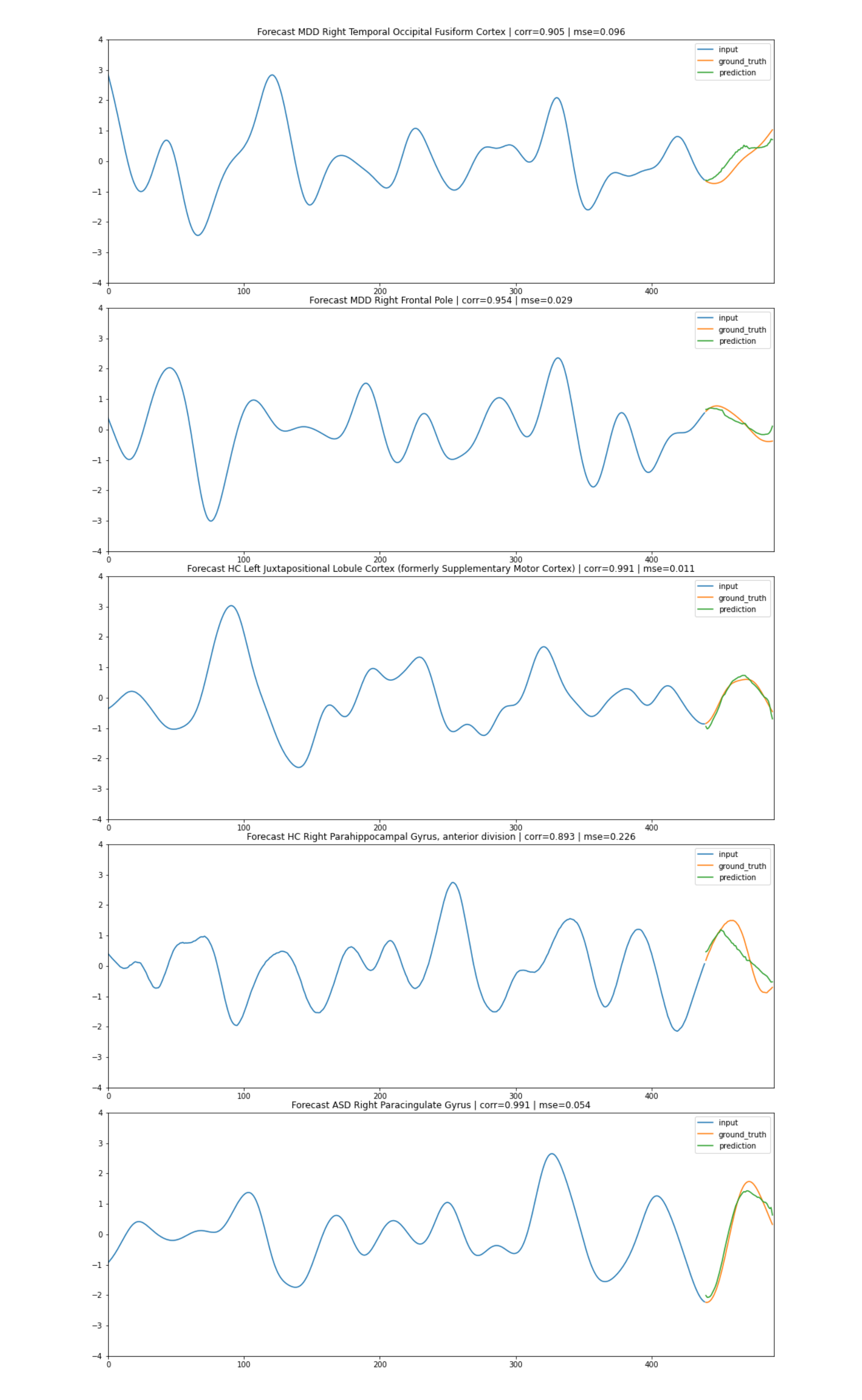}
\caption{Visualization of the Graph-S4 test-set predictions for forecasting $T$ = 50 timepoints as a self-supervised task on the healthy participants. The results show that the model can accurately forecast healthy and patients samples at inference time.} \label{fig3}
\end{figure}
 
 \newpage
\section{Datasets}

In this work we utilized three open-source multi-site clinical datasets. The  \textbf{Rest-Meta-MDD} consortium \cite{restmdd} does not share the raw 4D files for all the participants. Pre-processed Data \footnote{http://rfmri.org/REST-meta-MDD}  includes ROI timecourses for different brain atlases as well as 3D summary measures are provided. For exact details of the pre-processing we refer the reader to \cite{restmdd}. For \textbf{ABIDE I+II} and \textbf{SRPBS} raw data can be downloaded\footnote{$http://fcon_1000.projects.nitrc.org/indi/abide/databases.html$}\footnote{$https://bicr-resource.atr.jp/srpbsopen/$}. The preprocessing was done using the Configurable Pipeline for the Analysis of Connectomes (C-PAC)\footnote{$https://fcp-indi.github.io/docs/latest/user/quick.html$}. Our preprocessing pipeline, consisted of i) slice time and motion correction, ii) skull stripping,  iii) nuisance regression which included head motion, scanner drift, and physiological noise, iv)Band-pass filtering (0.01-0.1 Hz),  v) co-registration of the resulting rs-fMRI image to the subject's anatomical image. Finally, vi) the images were normalized onto the standard Montreal Neurological Institute (MNI) space (4 mm). For the \textbf{UkBioBank}\cite{ukbiobank}, the pre-processed 4D volumes in MNI space were provided. We used the Harvard-Oxford \cite{ho} atlas for all the datasets to segment and extract ROI timecourses. We excluded all participants with scanning duration less than 4 minutes and all the scanning sites with less than 10 participants from each group. We also excluded participants with low QC scores provided by the consortia if available. Information about the selected participants in this study is reported in Table \ref{table5}. To facilitate transfer learning from the \textbf{UkBioBank} to the clinical datasets, we resized the timecourses of the clinical datasets by bi-linear interpolation to approximately match their temporal revolution. We experimented with down-sampling the \textbf{UkBioBank} dataset and fixing the size and resolution of the clinical datasets, however we empirically found improved results with resizing the clinical datasets.

\begin{table}[]
\caption{Information about datasets and participants used in this study. $H$: Healthy,  $P$: Patients, $M$: Males, $F$: Females,  $TR$: Temporal resolution of the scan.  }\label{table6}
\centering
\begin{tabular}{|l|l|l|l|ll|l|}
\hline
              & N ($H$/$P$)        & $M$/$F$       & Age   (years)         & \multicolumn{1}{l|}{\# Sites} & $TR$  (sec.)          & \# Timepoints \\ \hline
Rest-Meta-MDD & 1453 (628/825)  & 546/907   & 38.1 $\pm$ 15.2 & \multicolumn{1}{l|}{11}       & 2.2 $\pm$  0.3 & 214 $\pm$ 20  \\ \hline
SRPBS         & 449 (264/185)   & 189//257  & 44.6 $\pm$ 13.5 & \multicolumn{1}{l|}{4}        & 2.2 $\pm$ 0.2  & 171 $\pm$ 54  \\ \hline
ABIDE I+II    & 1207 (558/649)  & 961/246   & 13.8 $\pm$ 6.2  & \multicolumn{1}{l|}{9}        & 2.0 $\pm$ 0.3  & 197 $\pm$ 40  \\ \hline
UKBioBank     & 19000 (19000/0) & 9500/9500 & 54 $\pm$ 7.5    & 3                             & 0.7            & 490           \\ \hline
\end{tabular}
\end{table}

\textit{All the consortia reported that the participants provided written informed consent. All recruitment procedures and experimental protocols were approved by the institutional review boards of their respective institutions.}

\section{Network prediction on the clinical datasets}

\begin{table}[h]
\caption{Average MSE of predicting the 8 networks on the three clinical datasets on the validation dataset.}\label{table7}
\centering
\begin{tabular}{|l|l|l|l|}
\hline
         & Rest-Meta-MDD & SRPBS & ABIDE I+II \\ \hline
Wavenet  & 0.288         & 0.232 & 0.309      \\ \hline
Informer & 0.451         & 0.475 & 0.491      \\ \hline
GWnet    & 0.294         & 0.310 & 0.350      \\ \hline
S4       & 0.289         & 0.208 & 0.346      \\ \hline
Graph-S4 & 0.273         & 0.201 & 0.322      \\ \hline
\end{tabular}
\end{table}

\section{Baselines Details}

\subsection{Self-supervised Baselines}
 
 \begin{itemize}
\item  \textbf{\textit{Wavenet}} We used 4 layers of dilated causal 1D CNNs with a kernel size = 3, dilation factors of 1,2,4,4 for each layer respectively and a feature dimension of 256 for all layers except the last layer with feature dimension equal to the number of ROIs of the target brain network. 
 \item \textbf{\textit{Informer}} We used 4 encoder and 2 decoder layers with 4 multi-head attention channels and a 512 as the number of features in the hidden layers. We used a sinusoidal function for positional encoding of the signal. All the other parameters were set to default setting recommended by the authors

 \item \textbf{\textit{GWnet}} We used 4 GWnet blocks (1D dilated causal CNNs with kernel size of 3 and a feature dimension of 32 followed by GCN layer). All the other parameters were set to default.
 
 \item \textbf{\textit{S4}} We used the same configuration as Graph-S4.
  \end{itemize}

 The baselines and Graph-S4 parameters were selected based on 30 runs of random hyperpramaeter search on the Ukbiobank dataset. We used a fixed training procedure for all the self-supervised for simplicity as described in the experiments section.

\subsection{Supervised Baselines}
 \begin{itemize}
\item  \textbf{\textit{SVM-linear \& SVM-rbf}} We used the SVM class in scikit-learn \footnote{https://scikit-learn.org/stable/modules/svm.html} library to train the models. We used $C$ as a hyper-parameter for the linear kernel  and $C$ and $\gamma$ as hyper-parameters for the radial basis function kernels and they were selected using a search on a inner loop. 
 \item \textbf{\textit{BrainNetCNN}} We used the configuration recommended by the author of 2 Edge-to-Edge Convolutions followed by two Convolutional layers and followed by 3 Fully connected layers with LeakRelLU activation function between all the layers. We replicated a python version of the publicly available code at \footnote{https://github.com/AmineEchraibi/BrainCNN/blob/master/BrainNetCNN.ipynb}

\item \textbf{\textit{GIN}} We used 3 layers of graph isomorphic convolutional layers with hidden feature dimension = 1024 and graph readout using concatenation of the summation of the embedded node features  followed by a fully connected layer for classification similar to \cite{kim2020understanding}. We used the correlation matrix as the adjacency matrix of the graphs. 

 \item \textbf{\textit{1D-CNNs}} We used 3 1D-CNN layers for feature extraction with a kernel size of 3 and feature dimension of 256. The rest of the parameters were set to default. 
 
 \item \textbf{\textit{ST-GCN}} We used the default configuration as provided by the open-source code \footnote{$https://github.com/sgadgil6/cnslab_fmri$}. 
 
  \item \textbf{\textit{DAST-GCN}} We used the default configuration as provided by the open-source code \footnote{https://github.com/AhmedElGazzar0/DAST-GCN}.

  \end{itemize}
\end{document}